\pdfoutput=1

\documentclass[11pt]{article}

\usepackage[review]{acl}

\usepackage{times}
\usepackage{latexsym}

\usepackage[T1]{fontenc}

\usepackage[utf8]{inputenc}

\usepackage{microtype}

\title{Extra Global Attention Designation Using Keyword Detection in Sparse Transformer Architectures}

\author{Evan Lucas \and Dylan Kangas \and Timothy C Havens\\
  Michigan Technological University / 1400 Townsend Drive \\
  Houghton, Michigan, United States of America \\
  
}

\begin{document}
\maketitle
\begin{abstract}
Transformers are deep network architectures that utilize a mechanism known as attention to create encoders, decoders, and encoder-decoder architectures. These architectures are, to date, producing leading performance in several machine learning tasks from natural language processing to computer vision. In this paper, we propose an extension to a popular sparse transformer architecture, called Longformer Encoder-Decoder.  One challenge with sparse transformers is that they can struggle with encoding of long range context, such as connections between topics discussed at a beginning and end of a document. A method to selectively increase global attention is proposed and demonstrated for summarization tasks on several benchmark data sets. By prefixing the transcript with additional keywords and encoding global attention on these keywords, improvement in zero-shot, few-shot, and fine-tuned cases is demonstrated for some benchmark data sets.
\end{abstract}

\section{Introduction}\label{sec:intro}

Early work in summarization was largely extractive, where key sentences or phrases from a document or data set were identified and the containing sentences were collected into a summary \cite{mihalcea2004textrank,riedhammer2008keyphrase}; that is, summaries were composed of phrases \emph{extracted} from the data. For applications such as dialogue summarization,  these purely extractive summarizations suffer from low readability, even if they capture the key parts of the text. Murray et. al \cite{murray2005extractive} introduce extractive oracles, a greedy algorithm with full knowledge of the human summary to select sentences to use in a machine generated summary. This concept of an extractive oracle has been used to demonstrate that an upper limit to the performance of extractive methods exists on several data sets \cite{hirao2017enumeration,zhong2021qmsum}, and is below the performance of many abstractive summarization methods. In contrast, abstractive summarization methods, where the information is rewritten into a more concise format and the summary does not necessarily use only words and phrases from document or transcript, have become more popular in the field of text summarization  \cite{banerjee2015abstractive}. 

The use of transformers, introduced in 2017 by Vaswani et al.~\cite{vaswani2017attention}, for summarization tasks has been studied in substantial depth \cite{feng2021survey}. After their introduction in 2017, they have quickly become one of the most popular methods for many \emph{natural language processing} (NLP) tasks including summarization. 
Transformers have many advantages over previously popular methods for working with sequential data. Namely, they do not require recurrence or convolutional layers and are able to process inputs as a whole. The transformer uses the concept of attention, which in loose terms can be thought of as similar to human attention; i.e., identifying parts of the input that are interesting and the connection between them.

Early transformers, such as the original transformer \cite{vaswani2017attention}, BERT \cite{kenton2019bert}, and BART \cite{lewis2020bart} were severely limited by input lengths, often 512 or 1024 tokens, due to the $O(n^2)$ time and space complexity required by the self-attention mechanism in the standard transformer. Several modifications to the transformer have been proposed to work around this, with one popular family of strategies using a sparse self-attention where many of the input pairs are selectively ignored. Some examples of this family include Big Bird \cite{zaheer2020big}, which uses a sliding window attention combined with randomly selected global attention; Longformer \cite{beltagy2020longformer}, which uses a sliding window attention and a dilated sliding window attention, along with some global attention tokens; and Smart Bird \cite{wu2021smart}, which uses a low-dimension full attention transformer to determine which attentions to calculate in the full-depth transformer. A representation of these different sparsity patterns is depicted in Figure \ref{globalAttnFigure} for an original transformer, Longformer, Big Bird, and our modifications to Longformer. The diagrams are meant to be illustrative only and exact numbers of cells may not be proportional to the actual sparse attention matrices used in each case.

One area of concern with these sparse transformer methods, as highlighted by \cite{tay2020long}, is that they can lack long range context. The tasks laid out in \cite{tay2020long} are varied and substantially different from summarization, but the concept is still similar: if relevant information is contained at a longer distance (or lags) in the input sequence than the sliding window, the model will not be able to compute the attention between those inputs. Other NLP specific tasks that require longer range context have also been proposed, such as LAMBADA, which tries to predict a word given context from sentences preceding the sentence missing the word. \cite{paperno2016lambada}

The original Longformer method includes global attention, but for summarization tasks the demonstrated method involves only adding global attention to the first  token, which is a special token that indicates to the model that it is a summarization task. Utilizing additional global attention has been proposed for the problem of multi-document summarization, where it helps provide long range context between documents using a modified Longformer \cite{caciularu2021cdlm}.  In this work, the leading sentences of each document are given global attention to provide context between documents.

The rest of this paper is organized as follows. The development, motivations, and evaluation methods selected for the method presented in this paper is presented in Section \ref{sec:methods}. 
Results of the method as applied to multiple long summarization data sets are presented in Section \ref{sec:results}. Results in Section \ref{sec:results} are divided into three sections; the zero and few shot case where training data is limited, the case where a full data set is available, and a small ablation study that attempts to understand the influence of some of the decisions used to create the method. 
We summarize in Section \ref{sec:conc}.

\section{Method}
\label{sec:methods}

In order to understand the modifications described later in this section, a short introduction to transformers is necessary. For a more thorough explanation of transformers, please see "Attention is all you need" \cite{vaswani2017attention}. For additional information on the modifications to the standard transformer used by Longformer, please see the Longformer paper \cite{beltagy2020longformer}. 

For emph{sequence-to-sequence} (seq2seq) tasks, such as summarization, where both the inputs and outputs of the model are sequences; a transformer is often used in the encoder-decoder structure. The sequence of input words is first tokenized, where a dictionary is used to replace words with tokens. Words and symbols appearing in the dictionary are replaced with a single token and other words are constructed using multiple tokens as needed. These tokens are given embeddings, which are a high dimensional learned representation of the token. A positional embedding to represent location within the input sequence is added directly to the token/word embedding.

This embedded representation of the sequence goes into the encoder of the transformer, where it goes through several layers of the same basic network. Each layer starts with a self attention computation, where attention is computed between the input sequence and itself. Attention, which is short for scaled dot product attention in this work and others, multiplies the embeddings by learned transformation matrices and computes a dot product (which is subsequently scaled and multiplied by yet another transformed embedding) between each pair of tokens considered. For the original transformer \cite{vaswani2017attention}, this is computed between each pair possible in the input sequence. For Longformer, some tokens are given "global attention" and attention is computed between that token and every other token in the sequence; however most of the tokens only have attention computed with tokens falling within a sliding window. This effectively sparsifies the attention matrix without losing local context. For traditional Longformer usage in summarization, global attention is also computed between the special token indicating that this is a summarization task. In our modification to Longformer, attention is also computed globally for some specifically selected tokens that have prefixed to the input.

Once attention is computed in the encoder layer, a residual connection from the embeddings adds the embeddings to the attention output; this is then normalized. A fully connected layer follows, with it's own residual connection and subsequent normalization. This overall process of computing attention and a fully connected layer is repeated for each encoder layer for a total of twelve encoder layers. When the "base" Longformer is used in place of the "large" Longformer model, there are only six encoder layers. 

This encoded representation flows into the decoder side of the encoder-decoder transformer. The decoder side starts by applying learned token and position embeddings to previously generated output tokens, then computes attention between these tokens and the encoded representation of the input. This is known as cross-attention. This cross attention goes into a self-attention computation, with a residual connection (followed by normalization) from the previously generated output. The output of the self attention has the typical residual connection and normalization, followed by a fully connected layer. Again, this entire set of actions is repeated for each decoder layer for a total of twelve layers. As with the previous comment about "large" and "base" Longformer models, the "base" model has six decoder layers.

After the final decoder layer, there is a single linear layer and a softmax. The output of this softmax is a distribution across the entire token distribution, showing the probability of each token being the next. Beam search is used to select sequences of tokens by considering multiple hypothesis sequences. This provides better results than a simple greedy search that selects the most probable next token at each step during summary generation and reduces the randomness introduced in sampling based approaches. In NLP tasks such as summarization and translation, beam search is often used for these reasons. \cite{yang2018breaking} During inference, each generated token is carried down to the beginning of the decoder layer so that the next token can be generated.  During training, teacher forcing \cite{lamb2016professor} is used, where masked versions of the reference output sequence are provided and only a single next token is computed from all prior tokens for each token in the input sequence. This method allows the parallel computation of many output tokens simultaneously.

Following other papers in this area, ROUGE scores \cite{lin2004rouge} are used to evaluate generated summaries.  For brevity, only the F-measure of the ROUGE scores are reported in our paper. The original formulation of the ROUGE scores are used, as implemented in the open source package released by Google Research and utilized by HuggingFace's Datasets library \cite{lhoest-etal-2021-datasets}. The version of ROUGE-L reported in this work is the original, and not the ROUGE-LSum; which is segmented at sentence boundaries or line breaks and is reported by some works such as \cite{see2017get}.

\emph{Longformer Encoder-Decoder} (LED) models were loaded and modified using the Huggingface Transformers \cite{wolf2020transformers} and PyTorch \cite{paszke2019pytorch} libraries. Data sets used for evaluation included the AMI meeting corpus \cite{mccowan2005ami} and the ICSI meeting corpus \cite{janin2003icsi}; the ConvoSumm benchmark data set \cite{fabbri2021convosumm} provided a pre-processed textual version of both of these. The AMI meeting corpus consists of sets of synthetic product design meetings that are acted out by several different teams. The ICSI meetings come from real student meetings at the Berkeley International Computer Science Institute and each meeting covers a variety of topics. The arXiv summarization data set, introduced by Cohan et al.~\cite{cohan2018discourse}, was also used. This data set consists of arXiv papers and their corresponding abstracts. All three of these data sets have substantially long inputs that would typically require truncation or chunking of the input if a sparse transformer was not used. 

\begin{figure}
        \centering
        \begin{subfigure}[b]{0.23\textwidth}
            \centering
            \includegraphics[width=\textwidth]{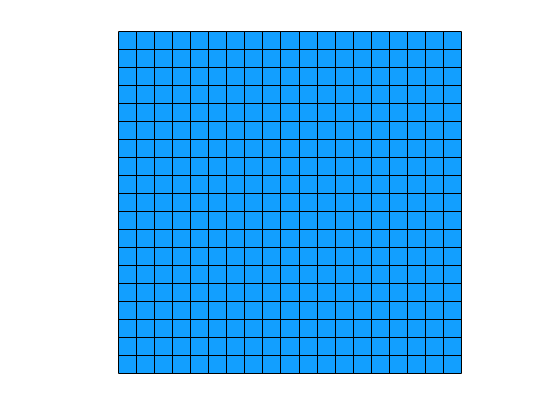}
            \caption[Network2]%
            {{\small Full $O(n^2)$ attention}}    
        \end{subfigure}
        \hfill
        \begin{subfigure}[b]{0.23\textwidth}  
            \centering 
            \includegraphics[width=\textwidth]{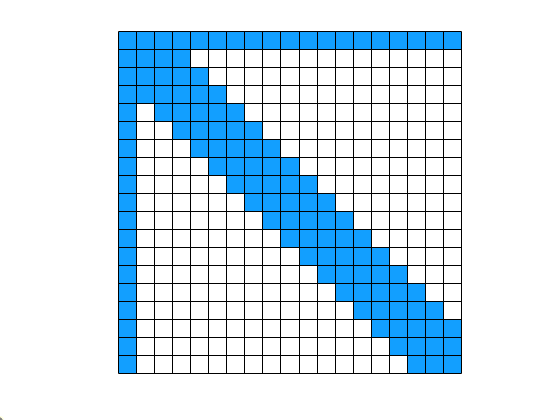}
            \caption[]%
            {{\small Longformer}}    
        \end{subfigure}
        \vskip\baselineskip
        \begin{subfigure}[b]{0.23\textwidth}   
            \centering 
            \includegraphics[width=\textwidth]{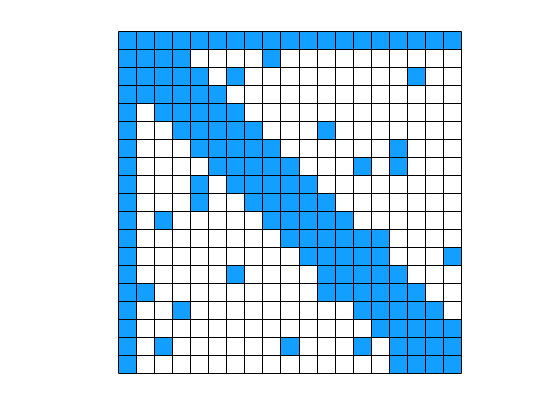}
            \caption[]%
            {{\small Big Bird}}    
        \end{subfigure}
        \hfill
        \begin{subfigure}[b]{0.23\textwidth}   
            \centering 
            \includegraphics[width=\textwidth]{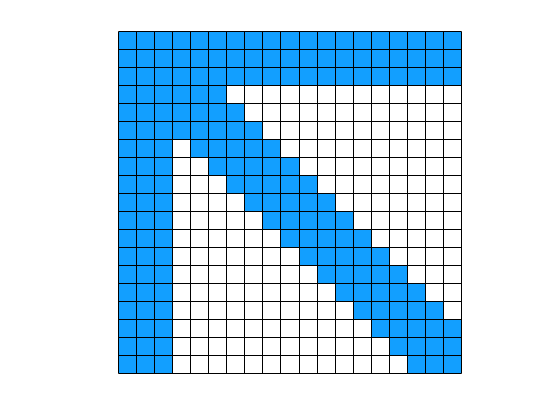}
            \caption[]%
            {{\small EGAD}}    
        \end{subfigure}
        \caption[]
        {Visualization of self-attention matrix sparsity. Blue filled cells represent attention being computed for that pair of inputs.} 
        \label{globalAttnFigure}
    \end{figure}

\begin{figure}
\centering
\includegraphics[width=8.5 cm]{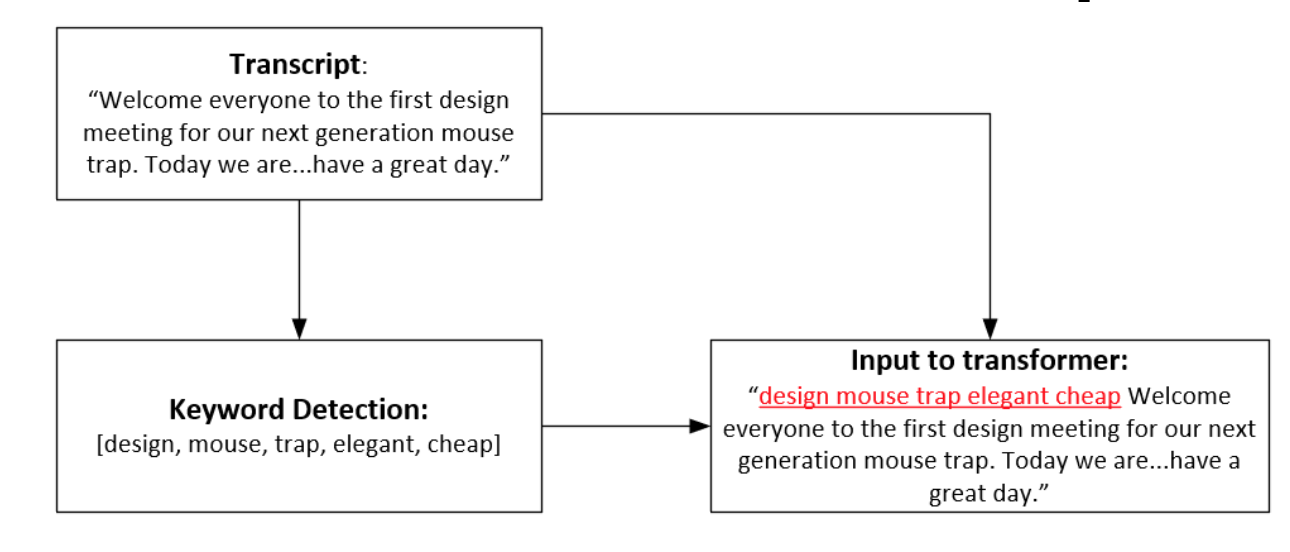}
\caption{Keyword prefixing block diagram. Keywords in red are given global attention (shown as blue rows/columns in Fig.~\ref{globalAttnFigure}.\label{blockDiagram}}
\end{figure}

This work attempts to improve the long range context by adding extra global attention tokens at the beginning of the input sequence. The base Longformer architecture of sliding window attention is used. Inspired in part by the concept of prefix tuning \cite{li2021prefix}, extra tokens that help maintain a topical focus are added to the beginning of the input and designated with global attention. A simple graphical example of this is depicted in Figure \ref{globalAttnFigure} (d). Following the method of Longformer, only the self attention for the encoder side is sparse and includes this extra global attention. An example of the workflow is depicted in Figure \ref{blockDiagram}, showing how the transcript is passed into the keyword detection method before the detected keywords are prefixed onto the transcript.


For the purposes of automated testing, the \emph{term frequency - inverse document frequency} (TF-IDF) method was used to select keywords that were disproportionately represented in the input sequence. Some real world applications would also be able to use human selected keywords or have prior knowledge from which to draw keywords. A dictionary of common English usage was used as the inverse dictionary to normalize word counts in the input text rather than have to compute the inverse document frequency for all of the possible input texts in a corpus. Initial investigations used Laplace smoothing \cite{hiemstra2009probability} to handle words not found in the dictionary; however, it was found that the very uncommon words not found in the dictionary did not improve results and unknown vocabulary was dropped from the keyword selection process.

To test whether the influence of this method came from having additional attention or the keywords themselves, random keyword selection was also tried. In order to constrain the random keywords to those that would exist in the text, they were uniformly sampled from the words present in each document. 

\section{Results}\label{sec:results}
Results are presented for two cases: the limited training data cases of zero-shot and few-shot, and the fine-tuned case where the body of training material is substantially larger. For all data sets, the training and validation splits used in the original data releases were sampled from for training and validation. Because this method is a modification of the LED model, a fine-tuned LED model without modification is used as a comparison for the presented method. For each data set considered, the model was fine-tuned for five epochs, and evaluated at each epoch. The model from each of these evaluations with the lowest validation loss was selected as the fine-tuned model.

\subsection{Zero and few-shot evaluation}
Based on the method used by \cite{xiao2021primer}, few-shot evaluations are performed on corpuses of 10 and 100 randomly selected training and evaluation examples. Five runs were performed with each configuration and averaged. To provide a comparison between different numbers of keywords, a pseudorandom seed was set so that each method used the same set of randomly chosen training data samples. For data sets with fewer than 100 samples in either the training or evaluation sets, the full set was used where appropriate. 

The first data set considered was the arXiv summarization corpus, which is presented in Table \ref{arXivFewShot}. The strongest improvements can be seen in the few-shot case with ten training samples. The improvement observed between base model and modified model is smaller to that observed by PRIMER, but still a very promising result. As the training corpus increases to 100 samples, there is a small detrimental effect with 10 keywords and only the 20 keyword case shows an improvement. In the zero-shot case, the extra attention appears to have a detrimental effect.

\begin{table}
\centering
\caption{Zero and few shot results with arXiv data set\label{arXivFewShot}}
\begin{tabular}{|p{1.0cm}|p{1.0cm}|p{1.0cm}|p{1.0cm}|p{1.0cm}|}
\hline
\textbf{Training Samples}	&\textbf{Model}	& \textbf{ROUGE-1}	& \textbf{ROUGE-2} & \textbf{ROUGE-L}\\
\hline
\multirow{4}{*}{0} 
                                & \multicolumn{1}{l|}{LED \cite{xiao2021primer}} & \multicolumn{1}{l|}{13.9} & \multicolumn{1}{l|}{3.8} & \multicolumn{1}{l|}{8.4} 
                                \\\cline{2-5}
                                & \multicolumn{1}{l|}{PRIMER \cite{xiao2021primer}} & \multicolumn{1}{l|}{29.1} & \multicolumn{1}{l|}{8.6} & \multicolumn{1}{l|}{15.8} 
                                 \\\cline{2-5}
                                & \multicolumn{1}{l|}{LED (ours)} & \multicolumn{1}{l|}{14.4} & \multicolumn{1}{l|}{3.4} & \multicolumn{1}{l|}{9.4} \\\cline{2-5}
                                 & \multicolumn{1}{l|}{10 keywords} & \multicolumn{1}{l|}{10.0} & \multicolumn{1}{l|}{2.2} & \multicolumn{1}{l|}{6.6}  \\\cline{2-5}
                                 & \multicolumn{1}{l|}{20 keywords} & \multicolumn{1}{l|}{8.1} & \multicolumn{1}{l|}{2.0} & \multicolumn{1}{l|}{5.5}

                                 \\\hline
\multirow{4}{*}{10} 
                                & \multicolumn{1}{l|}{LED \cite{xiao2021primer}} & \multicolumn{1}{l|}{36.5} & \multicolumn{1}{l|}{11.2} & \multicolumn{1}{l|}{20.7} 
                                \\\cline{2-5}
                                & \multicolumn{1}{l|}{PRIMER \cite{xiao2021primer}} & \multicolumn{1}{l|}{41.1} & \multicolumn{1}{l|}{13.8} & \multicolumn{1}{l|}{23.0} 
                                 \\\cline{2-5}
                                & \multicolumn{1}{l|}{LED (ours)} & \multicolumn{1}{l|}{31.9} & \multicolumn{1}{l|}{11.6} & \multicolumn{1}{l|}{16.9} \\\cline{2-5}
                                 & \multicolumn{1}{l|}{10 keywords} & \multicolumn{1}{l|}{34.5} & \multicolumn{1}{l|}{12.7} & \multicolumn{1}{l|}{17.9}  \\\cline{2-5}
                                 & \multicolumn{1}{l|}{20 keywords} & \multicolumn{1}{l|}{33.9} & \multicolumn{1}{l|}{12.9} & \multicolumn{1}{l|}{17.9} 
                               
                                \\\hline
  \multirow{4}{*}{100} 
                                & \multicolumn{1}{l|}{LED \cite{xiao2021primer}} & \multicolumn{1}{l|}{41.0} & \multicolumn{1}{l|}{13.7} & \multicolumn{1}{l|}{22.3} 
                                \\\cline{2-5}
                                & \multicolumn{1}{l|}{PRIMER \cite{xiao2021primer}} & \multicolumn{1}{l|}{43.4} & \multicolumn{1}{l|}{15.9} & \multicolumn{1}{l|}{24.1} 
                                 \\\cline{2-5}
                                & \multicolumn{1}{l|}{LED (ours)} & \multicolumn{1}{l|}{36.8} & \multicolumn{1}{l|}{13.8} & \multicolumn{1}{l|}{19.8} \\\cline{2-5}
                                 & \multicolumn{1}{l|}{10 keywords} & \multicolumn{1}{l|}{36.7} & \multicolumn{1}{l|}{13.9} & \multicolumn{1}{l|}{19.8}  \\\cline{2-5}
                                 & \multicolumn{1}{l|}{20 keywords} & \multicolumn{1}{l|}{37.7} & \multicolumn{1}{l|}{14.3} & \multicolumn{1}{l|}{20.4}
                               
                                \\\hline                               
\end{tabular}
\end{table}

The AMI and ICSI data sets were also considered and results of each are presented in Tables \ref{AMIFewShot} and \ref{ICSIFewShot}, respectively. Using the AMI data, small improvements were observed in the zero-shot case, along with small improvements for most of the few-shot cases. Interestingly, when considering the ICSI data, the addition of extra attention did not improve scores in most cases. Only the 10-sample few-shot case showed improved scores when extra attention was added. It is hypothesized that this is because the ICSI transcripts and summaries span multiple topics and the added keywords and global attention cause a focus on only some of the topics. 

\begin{table}
\centering
\caption{Zero and few shot results with AMI data set\label{AMIFewShot}}
\begin{tabular}{|p{1.0cm}|p{1.0cm}|p{1.0cm}|p{1.0cm}|p{1.0cm}|}
\hline
\textbf{Training Samples}	&\textbf{Keyword Count}	& \textbf{ROUGE-1}	& \textbf{ROUGE-2} & \textbf{ROUGE-L}\\
\hline
\multirow{4}{*}{0} & \multicolumn{1}{l|}{0} & \multicolumn{1}{l|}{19.2} & \multicolumn{1}{l|}{3.2} & \multicolumn{1}{l|}{9.8} \\\cline{2-5}
                                 & \multicolumn{1}{l|}{10} & \multicolumn{1}{l|}{21.9} & \multicolumn{1}{l|}{3.3} & \multicolumn{1}{l|}{10.7}  \\\cline{2-5}
                                 & \multicolumn{1}{l|}{20} & \multicolumn{1}{l|}{20.7} & \multicolumn{1}{l|}{3.9} & \multicolumn{1}{l|}{9.9}  \\\hline
\multirow{4}{*}{10} & \multicolumn{1}{l|}{0} & \multicolumn{1}{l|}{34.5} & \multicolumn{1}{l|}{12.3} & \multicolumn{1}{l|}{18.5} \\\cline{2-5}
                                 & \multicolumn{1}{l|}{10} & \multicolumn{1}{l|}{34.8} & \multicolumn{1}{l|}{12.1} & \multicolumn{1}{l|}{19.0}  \\\cline{2-5}
                                 & \multicolumn{1}{l|}{20} & \multicolumn{1}{l|}{33.3} & \multicolumn{1}{l|}{11.8} & \multicolumn{1}{l|}{18.6}  \\\hline
  \multirow{4}{*}{100} & \multicolumn{1}{l|}{0} & \multicolumn{1}{l|}{44.4} & \multicolumn{1}{l|}{15.5} & \multicolumn{1}{l|}{21.9} \\\cline{2-5}
                                 & \multicolumn{1}{l|}{10} & \multicolumn{1}{l|}{45.7} & \multicolumn{1}{l|}{16.1} & \multicolumn{1}{l|}{22.4}  \\\cline{2-5}
                                 & \multicolumn{1}{l|}{20} & \multicolumn{1}{l|}{46.0} & \multicolumn{1}{l|}{16.6} & \multicolumn{1}{l|}{22.6}  \\\hline                               
\end{tabular}
\end{table}

\begin{table}
\centering
\caption{Zero and few shot results with ICSI data set\label{ICSIFewShot}}
\begin{tabular}{|p{1.0cm}|p{1.0cm}|p{1.0cm}|p{1.0cm}|p{1.0cm}|}
\hline
\textbf{Training Samples}	&\textbf{Keyword Count}	& \textbf{ROUGE-1}	& \textbf{ROUGE-2} & \textbf{ROUGE-L}\\
\hline
\multirow{4}{*}{0} & \multicolumn{1}{l|}{0} & \multicolumn{1}{l|}{18.0} & \multicolumn{1}{l|}{1.8} & \multicolumn{1}{l|}{8.7} \\\cline{2-5}
                                 & \multicolumn{1}{l|}{10} & \multicolumn{1}{l|}{15.2} & \multicolumn{1}{l|}{1.5} & \multicolumn{1}{l|}{6.3}  \\\cline{2-5}
                                 & \multicolumn{1}{l|}{20} & \multicolumn{1}{l|}{9.8} & \multicolumn{1}{l|}{1.1} & \multicolumn{1}{l|}{4.2}  \\\hline
\multirow{4}{*}{10} & \multicolumn{1}{l|}{0} & \multicolumn{1}{l|}{28.5} & \multicolumn{1}{l|}{7.5} & \multicolumn{1}{l|}{13.7} \\\cline{2-5}
                                 & \multicolumn{1}{l|}{10} & \multicolumn{1}{l|}{30.0} & \multicolumn{1}{l|}{8.6} & \multicolumn{1}{l|}{14.3}  \\\cline{2-5}
                                 & \multicolumn{1}{l|}{20} & \multicolumn{1}{l|}{30.0} & \multicolumn{1}{l|}{8.1} & \multicolumn{1}{l|}{13.7}  \\\hline
  \multirow{4}{*}{100} & \multicolumn{1}{l|}{0} & \multicolumn{1}{l|}{40.1} & \multicolumn{1}{l|}{10.7} & \multicolumn{1}{l|}{18.1} \\\cline{2-5}
                                 & \multicolumn{1}{l|}{10} & \multicolumn{1}{l|}{39.5} & \multicolumn{1}{l|}{11.3} & \multicolumn{1}{l|}{17.9}  \\\cline{2-5}
                                 & \multicolumn{1}{l|}{20} & \multicolumn{1}{l|}{38.9} & \multicolumn{1}{l|}{11.3} & \multicolumn{1}{l|}{17.9}  \\\hline                               
\end{tabular}
\end{table}

\subsection{Full set evaluation results}
The usefulness of the proposed method appears to be data specific. In particular, the benefit of adding extra global attention is strongest when the corpus is small, as demonstrated in the few and zero shot evaluation section. The model also has to be pre-trained with the extra keywords present for the largest benefit. In large data sets, such as the arXiv summarization data set \cite{cohan2018discourse}, using a model pre-trained on the data set showed no improvements when extra keywords were added. No benefit was also found when attempting to fine-tune a model with extra attention that was already well trained on this data set.

To fully test the effect on a small (but complete) corpus, the AMI and ICSI data sets were considered in full. The results of the AMI data set are presented in Table \ref{AMIfineTune}. For reference, the current \emph{state of the art} (SOTA) results from \cite{fabbri2021convosumm} are presented. Although these (previously published) results could not be reproduced by our efforts, we demonstrate a larger step improvement in scores by the addition of extra attention than was demonstrated in the SOTA paper by the addition of argument-mining. 

The full ICSI corpus results are presented in Table \ref{ICSIfineTune}. Again, results from the current SOTA paper are shared. As discussed above, the wide range of topics discussed in each transcript and summary appear to disrupt the proposed method of adding attention with keywords, and the addition of extra attention degrades summarization performance.

\begin{table}
\begin{center}
\caption{Full set evaluation results with AMI data set\label{AMIfineTune}}
\begin{tabular}{|p{2.0cm}|p{1.7cm}|p{1.7cm}|p{1.7cm}|}
\hline
\textbf{Keyword Count}	& \textbf{ROUGE-1}	& \textbf{ROUGE-2} & \textbf{ROUGE-L}\\
\hline
Base LED (\cite{fabbri2021convosumm})*		& 54.20	& 20.72 & **\\
\hline
LED-arg (\cite{fabbri2021convosumm})		& 54.47	& 20.83 & **\\
\hline
Base LED (ours)*		& 46.1	& 17.4 & 23.1\\
\hline
10 keywords		& 46.6	& 16.9 & 23.1\\

\hline
20 keywords		& 46.7	& 17.0 & 24.1\\
\hline
50 keywords		& 48.5	& 17.7 & 23.7\\
\hline
100 keywords		& 48.4	& 17.4 & 23.3\\

\hline
\end{tabular}\end{center}
*Attempts to replicate these previously published results with author-provided code were unsuccessful. Author was contacted and offered some suggestions, but it did not resolve the difference in results.\\
**Comparable ROUGE-L scores from this work are not available due to differences in method of computing ROUGE-L.
\end{table}

\begin{table}
\begin{center}
\caption{Full set evaluation results with ICSI data set\label{ICSIfineTune}}
\begin{tabular}{|p{2.0cm}|p{1.7cm}|p{1.7cm}|p{1.7cm}|}
\hline
\textbf{Keyword Count}	& \textbf{ROUGE-1}	& \textbf{ROUGE-2} & \textbf{ROUGE-L}\\
\hline
Base LED (\cite{fabbri2021convosumm})		& 43.03	& 12.14 & **\\
\hline
LED-arg (\cite{fabbri2021convosumm})		& 44.17	& 11.09 & **\\
\hline
Base LED (ours)		& 45.5	& 13.3 & 19.2\\

\hline
10 keywords		& 40.0	& 11.8 & 18.1\\

\hline
20 keywords		& 40.9	& 11.8 & 18.0\\
\hline
50 keywords		& 38.9	& 12.0 & 17.7\\
\hline
100 keywords		& 38.8	& 11.9 & 17.0\\
\hline

\end{tabular}\end{center}
**Comparable ROUGE-L scores from this work are not available due to differences in method of computing ROUGE-L.
\end{table}

\subsection{Sensitivity study}
To understand the influence of the keywords used for the extra attention, a small sensitivity study was performed. Two other keyword selection methods were compared to the TF-IDF method: i) random keywords and ii) ``gibberish''. For these two additional methods, five sets of random keywords or gibberish were tested and average results are reported. Gibberish was generated by uniformly sampling the lowercase English alphabet using a word length sampled from a binomial distribution with a population of ten and a probability of 0.5. It was observed that some individual runs with random words provided better results than the those selected using TF-IDF, but overall the TF-IDF keyword detection method was superior. Results of these trials are shared in Table \ref{ablation}. As expected, random keywords and nonsense words degrade performance, with gibberish degrading performance by a larger margin.

\begin{table}
\centering
\caption{Sensitivity study\label{ablation}}
\begin{tabular}{|p{2.6cm}|p{1.4cm}|p{1.4cm}|p{1.4cm}|}
\hline
\textbf{Keyword Count}	& \textbf{ROUGE-1}	& \textbf{ROUGE-2} & \textbf{ROUGE-L}\\
\hline

\hline
Base LED 		& 46.1	& 17.4 & 23.1\\
\hline
10 keywords		& 46.6	& 16.9 & 23.1\\
\hline
10 random keywords		& 45.3	& 16.4 & 22.2\\

\hline
10 gibberish words		& 44.6	& 16.4 & 22.7\\

\hline

\end{tabular}

\end{table}

\section{Conclusions}\label{sec:conc}
A method for adding long range context to Longformer is proposed and tested. Improvements on few-shot and small datasets are demonstrated. The apparent main limitation of the method, degraded performance on multi-topic summaries, is also identified and discussed. 

One area for potential improvements with this method is to test improved keyword selection methods. TF-IDF is a fairly primitive method and more sophisticated options may provide better keywords that aren't present in the original text. The ability of the method to steer a summary could also be investigated in future work, although this would also require either subjective human assessment or a new method for measuring the influence of the keywords.

\subsection{References}

\nocite{Ando2005,borschinger-johnson-2011-particle,andrew2007scalable,rasooli-tetrault-2015,goodman-etal-2016-noise,harper-2014-learning}

The \LaTeX{} and Bib\TeX{} style files provided roughly follow the American Psychological Association format.
If your own bib file is named \texttt{custom.bib}, then placing the following before any appendices in your \LaTeX{} file will generate the references section for you:
\begin{quote}
\begin{verbatim}
\bibliographystyle{acl_natbib}
\bibliography{custom}
\end{verbatim}
\end{quote}

You can obtain the complete ACL Anthology as a Bib\TeX{} file from \url{https://aclweb.org/anthology/anthology.bib.gz}.
To include both the Anthology and your own .bib file, use the following instead of the above.
\begin{quote}
\begin{verbatim}
\bibliographystyle{acl_natbib}
\bibliography{anthology,custom}
\end{verbatim}
\end{quote}

Please see Section~\ref{sec:bibtex} for information on preparing Bib\TeX{} files.

\bibliography{anthology,references}
\bibliographystyle{acl_natbib}

\end{document}